\documentclass[runningheads]{llncs}

\usepackage{iciapabbrv}
\usepackage{iciap}

\usepackage{graphicx}
\usepackage{booktabs}

\usepackage[accsupp]{axessibility}  

\usepackage[pagebackref,breaklinks,colorlinks,citecolor=iciapblue]{hyperref}

\usepackage{orcidlink}

\begin{document}

\title{Hierarchical Vision-Language Retrieval of Educational Metaverse Content in Agriculture}

\titlerunning{Hierarchical V-L Retrieval of EduMetaverse Content in Agriculture}

\author{Ali Abdari\inst{1,2}\orcidlink{0000-0002-4482-0479} \and
Alex Falcon\inst{1}\orcidlink{0000-0002-6325-9066} \and
Giuseppe Serra\inst{1}\orcidlink{0000-0002-4269-4501}}

\authorrunning{A.~Abdari et al.}

\institute{University of Udine, Via delle scienze 206, Udine, 331100, Italy \and
University of Naples Federico II, C.so Umberto I, 40, Naples, Italy
\email{\{abdari.ali,falcon.alex\}@spes.uniud.it, giuseppe.serra@uniud.it}}

\maketitle

\begin{abstract}
Every day, a large amount of educational content is uploaded online across different areas, including agriculture and gardening. When these videos or materials are grouped meaningfully, they can make learning easier and more effective. One promising way to organize and enrich such content is through the Metaverse, which allows users to explore educational experiences in an interactive and immersive environment. However, searching for relevant Metaverse scenarios and finding those matching users' interests remains a challenging task. A first step in this direction has been done recently, but existing datasets are small and not sufficient for training advanced models. In this work, we make two main contributions: first, we introduce a new dataset containing 457 agricultural-themed virtual museums (AgriMuseums), each enriched with textual descriptions; and second, we propose a hierarchical vision-language model to represent and retrieve relevant AgriMuseums using natural language queries. In our experimental setting, the proposed method achieves up to about 62\% R@1 and 78\% MRR, confirming its effectiveness, and it also leads to improvements on existing benchmarks by up to 6\% R@1 and 11\% MRR. Moreover, an extensive evaluation validates our design choices. 
Code and dataset are available at \href{https://github.com/aliabdari/Agricultural_Metaverse_Retrieval}{https://github.com/aliabdari/Agricultural\_Metaverse\_Retrieval}.
  
\end{abstract}

\section{Introduction}
As user-generated content grows online, platforms like YouTube have become popular for sharing tutorials on everything from cooking and home repairs to programming and agriculture \cite{youtube_education}. Many videos offer step-by-step guidance on tasks like planting vegetables or setting up irrigation. However, the abundance of content can be overwhelming, especially for less informed users such as older adults or some farmers, and videos may lack complete information to solve specific problems. The metaverse offers a promising alternative, allowing users to engage with content more effectively. In virtual environments, related information can be gathered in one space, making it easier to explore topics comprehensively.

Recent initiatives, like the FARMVR framework \cite{FarmVR2024} offer interactive VR experiences that promote agricultural literacy. These projects highlight how metaverse technologies are transforming traditional learning by delivering engaging, hands-on experiences that improve both understanding and skill acquisition. One of the new tasks in this area is retrieving a suitable metaverse museum based on the given textual queries by the users like our previous work \cite{abdari2025agrimus}, in which we use virtual reality to help students and educators explore agricultural concepts without relying on physical resources. However, they lack a sufficient amount of agricultural museum content and also are relied on non learning zero shot approaches to retrieve suitable metaverse based on the textual queries. These defects limit the ability to cover a wide range of topics in the agricultural sector. Moreover, due to their reliance on zero-shot approaches, they are not robust in all conditions, for example, in distinguishing between similar content.

\begin{figure}
    \centering
    \includegraphics[width=\linewidth,trim=1.5cm 13cm 2cm 5cm,clip]{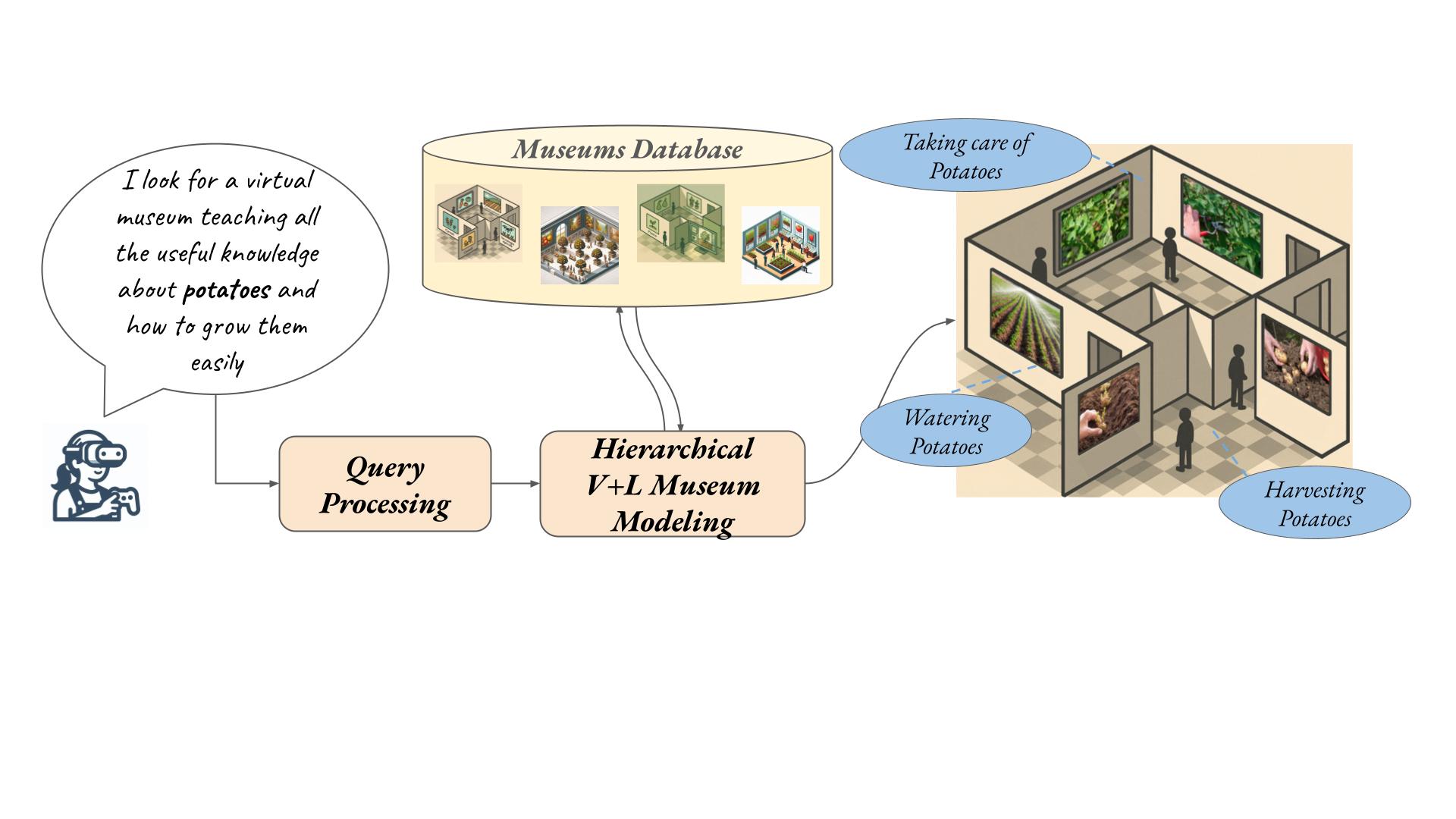}
    \caption{Overview of the task of Retrieval of thematic Agricultural Museums based on the given Textual Queries.}
    \label{fig:overview_task}
\end{figure}

Therefore, in order to leverage the capabilities of the metaverse in training the agricultural content, this work aims to create thematic museums in the agricultural sector, enabling users to find the most suitable environments based on their textual queries. To this end, we propose an approach to process the thematic museums and textual queries at the same time to enable final users to retrieve relevant museums based on natural language input. The overview of the defined task can be seen in Figure \ref{fig:overview_task}. 

Overall, the contributions of this work can be summarized as follows:
\begin{itemize} 
    \item We introduce a new dataset containing 457 thematic agricultural museums, enriched with descriptive information, as the currently available datasets are very small and lack descriptions, preventing the training of a retrieval system using advanced machine learning models and architectures. 
    \item We build a retrieval system to enable users to find their desired museum using textual queries by implementing a hierarchical vision-language approach to represent both the structure of the museums and the textual descriptions.
\end{itemize}

In the following sections, Section \ref{sec_related_work} presents related works in the area of analyzing 3D scenarios and virtual museums. The proposed approach for the retrieval task of finding Agricultural museums based on the textual queries is described in Section \ref{sec_method}. The created dataset and the experimental results are presented in Section \ref{sec_exp}. The limitations and potential directions for future work are discussed in Section \ref{sec_limitations}, and finally, Section \ref{sec_conclusion} summarizes the main contributions and findings of this study.

\section{Related Works \label{sec_related_work}}

\subsection{Virtual Exhibits Beyond Cultural Heritage}
The use of virtual exhibits has become more popular as a new way to support education and public access. Digital museums, which often focus on cultural heritage, now use immersive technologies like virtual reality to create interactive experiences that people can access from anywhere \cite{kiourt2016dynamus, zidianakis2021invisible}. These platforms help preserve artifacts and also let users explore content in an engaging way \cite{barszcz20233d, merella2023structured}.

While many of these platforms focus on historical or artistic collections, recent studies have begun to explore alternative domains. In particular, works like \cite{falcon2024hierartex} propose digital museums enriched with video content and structured layouts aimed at educational purposes. Our research expands on this idea by applying the digital museum format to agriculture-related knowledge, creating an immersive learning environment for topics such as home or open-field farming.

\subsection{Cross-modal Scene Understanding in 3D Environments}

The ability to link natural language with 3D scenarios has significantly advanced thanks to recent developments in vision-language learning. Systems such as Scan2Cap \cite{chen2021scan2cap} and TOD3Cap \cite{jin2024tod3cap} have demonstrated how dense captioning techniques can describe spatial arrangements in both indoor and outdoor environments, respectively. These models use relational reasoning and multi-modal fusion to generate descriptive text that mirrors the 3D layout of a scene.

Moreover, large-scale frameworks like PQ3D \cite{zhu2024unifying} and GPT4Scene \cite{qi2025gpt4scene} have made it possible to handle multiple 3D understanding tasks under a unified system. These tools are particularly relevant for applications that require retrieval, manipulation, or explanation of complex 3D spaces through text.

\subsection{Text-based Retrieval of Complex Multimedia-rich Scenes}

Recently, several works have focused on the retrieval of richly structured 3D scenarios using natural language queries. Unlike early approaches focused on basic object-level queries, newer methods address the complexity of multi-room layouts containing different objects \cite{yu2024towards,abdari2023farmare}. For example, CRISP \cite{yu2024towards} offers large datasets of 3D interiors with textual descriptions, and FArMARe \cite{abdari2023farmare} and AdOCTeRA \cite{abdari2024adoctera, abdari2025reproducibility} introduce methods to rank scenes based on relevance to user query.

However, existing methods often overlook the presence of embedded multimedia, such as images or videos, which can be crucial for interpreting scene content. Our recent works \cite{abdari2023metaverse, falcon2024hierartex, abdari2024language, falcon2024paving} have highlighted this gap, suggesting the integration of additional modalities to improve semantic alignment with user queries. We extend this by integrating video into 3D museums and improving multimedia retrieval.

\section{Proposed Method \label{sec_method}}
To effectively capture the structure of our dataset, comprising museums, rooms, and topic-based videos, we adopt a hierarchical learning framework that reflects this multi-level organization. 
By modeling the data in this way, each layer of the system learns representations considering dependencies within and between entities. For example, museum-level information can influence how rooms are organized, which in turn has information about the content of individual videos. This structure helps the system retrieve content not only based on video-level relevance but also with awareness of the museum's overall context and structure.

\begin{figure}[ht]
    \centering
    \includegraphics[width=\linewidth,trim=3cm 6.5cm 6cm 7cm,clip]{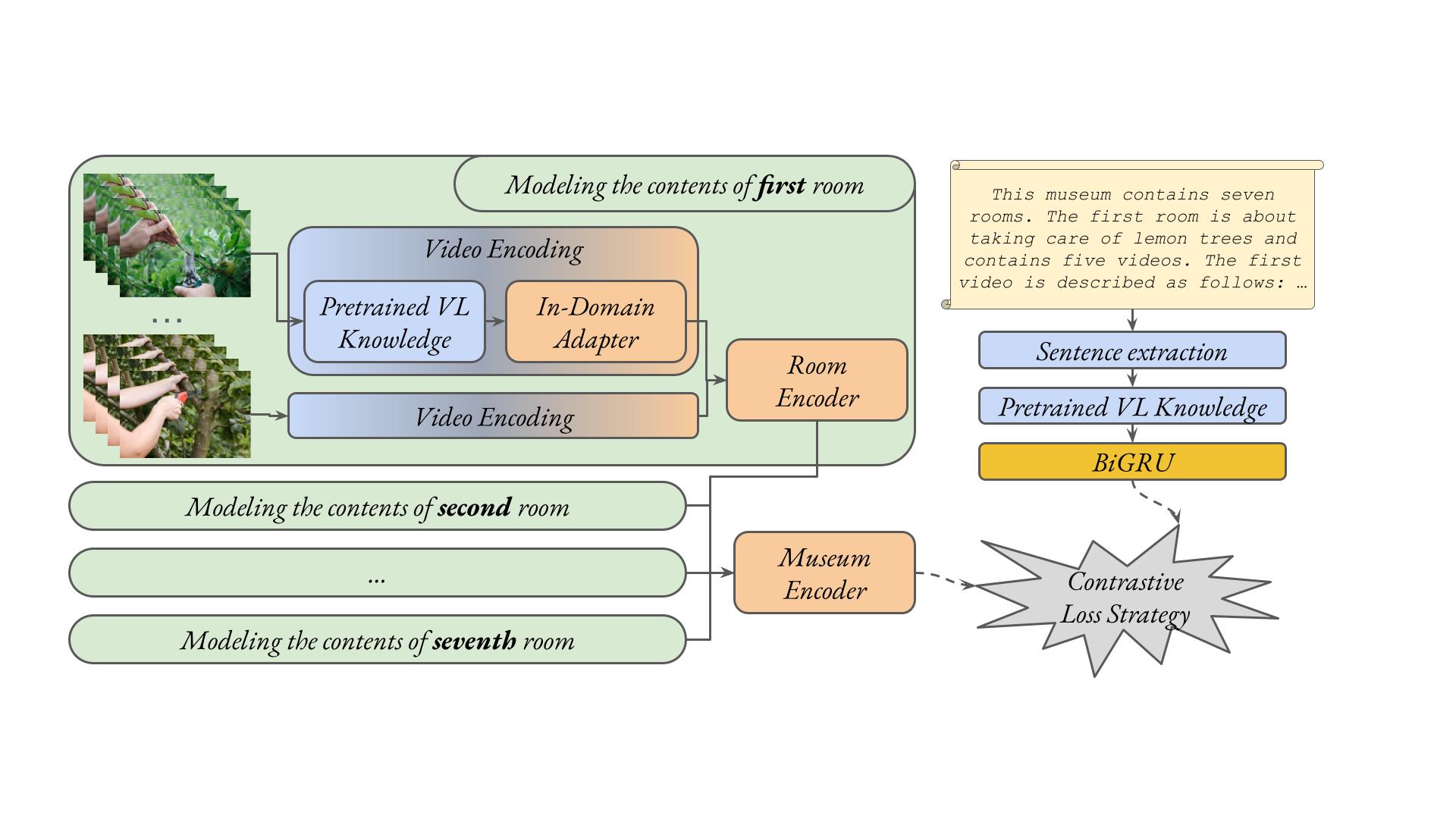}
    \caption{A detailed view of the proposed method. Details in Section \ref{sec_method}.}
    \label{fig:detail-arch}
\end{figure}

The proposed method is depicted in Figure \ref{fig:detail-arch}. Information in each museum is progressively processed and integrated in a bottom-up fashion. Specifically, the videos within each room are first encoded by the Video Encoding module, which comprises two components: Pretrained VL Knowledge and In-Domain Adapter. The Pretrained VL Knowledge component incorporates general representations derived from models pretrained on large-scale datasets (e.g., Clip \cite{radford2021learning}) and remains frozen during training. In contrast, the In-Domain Adapter is a lightweight module introduced to better adapt the frame representations extracted from the pretrained model to the target domain, while also learning to aggregate them. It consists of four layers: a 1D convolution (Conv1D), followed by Batch Normalization, ReLU activation, and a Linear projection.

The representations obtained from individual videos are then aggregated by the Room Encoder, which follows a similar architecture to the In-Domain Adapter. This module aims to integrate video-level information to obtain a comprehensive representation of each room. Subsequently, Museum Encoder, which mirrors the structure of the Room Encoder, aggregates room-level representations to build a global representation of the entire museum.

To enable text-based retrieval, two additional modules are introduced. The first processes textual information derived from each museum. This textual input is generated by filling a predefined template with structured information, such as the number of rooms (e.g., seven), the theme of each room (e.g., taking care of lemon trees), and the description of the individual videos. The resulting paragraph, which is typically long, is split into individual sentences. Each sentence is first encoded using the textual encoder from the Pretrained Knowledge module, and then processed by a Bidirectional GRU. This approach is inspired by prior work addressing similar challenges with long textual inputs \cite{abdari2023farmare,falcon2024hierartex}.

Finally, the entire system is trained using a contrastive learning objective. In particular, we adopt a triplet loss to align visual and textual representations in a shared embedding space \cite{schroff2015facenet}.



\section{Experimental Results \label{sec_exp}}
In our experiments, we aim to address six key research questions. First, what is the effect of replacing the default vision-language model with alternative architectures of a similar class? Second, considering the video-based nature of the input, what are the benefits of using large-scale video models? Third, does combining vision-language models with video-only models lead to improved performance compared to using either type in isolation? Fourth, how well do vision-language models perform in a zero-shot setting on our dataset? Fifth, how effective is the proposed hierarchical structure, and how does performance change when it is ablated? Finally, does training on the collected dataset lead to improved performance on existing, related benchmarks?

Before moving on to answering these questions, we introduce the newly collected dataset in Section \ref{subsec_dataset}, the evaluation metrics in Section \ref{sec:eval}, and the experimental settings in Section \ref{sec:config}.

\subsection{Collected dataset \label{subsec_dataset}}
To create thematic agricultural museums, we use an approach similar to our previous work \cite{abdari2025agrimus}. Differently, we consider a range of 3 to 8 rooms per museum to increase diversity, with 2 to 4 videos assigned to each room. On average, each museum contains 11.45 videos, 2.50 videos per room, and 4.57 rooms per museum.

Since the final goal is to create a system where users can find thematic Metaverses through text searches, we add detailed descriptions to each of the museums we created. Each description includes information about the number of rooms in the museum, the number of videos in each room, and details about each video. The descriptions start with general information about the number of rooms (for example: "This museum has six rooms"). Then, more specific details about each room are given (for example: "The first room has three videos"). Finally, each video is described using its title (for example: "The first video is about Indoor Vegetable Growing"). In this way, we provide a clear and complete description of all the available content in each museum. On average, each of the created descriptions contains 18.72 sentences and 242.85 tokens.

Overall, the collected dataset contains 457 agricultural thematic museums, of which 70\% (i.e., 319 museums) are used for training, while 15\% are allocated to each of the validation and test sets (i.e., 69 museums each). 

\subsection{Evaluation metrics \label{sec:eval}}
To evaluate the overall performance, we chose the most commonly used metrics in research fields related to cross-modal retrieval. \textbf{R@K} refers to the recall metric, where K can be 1, 5, or 10, and it indicates the percentage of cases in which the most relevant museum appears within the top K ranks. Additionally, \textbf{MedR} and \textbf{MeanR} represent the median and mean ranks, respectively, while \textbf{MRR} denotes the mean reciprocal rank.

\subsection{Configuration\label{sec:config}}
All experiments were conducted on a system with 16 GB of RAM, an Intel(R) Xeon(R) CPU E5-1620 v4 @ 3.50GHz, and an NVIDIA RTX A5000 GPU. The experiments were run using Python 3.10 and PyTorch version 2.5.0+cu124. In all the experiments, we used 0.0007 as the learning rate with the batch size of 64. The training was performed for up to 50 epochs, with early stopping applied if the validation loss did not improve for 25 consecutive epochs, using a delta of 0.0001. 
    
\subsection{Effect of vision-language model choice \label{subsec_vision_language_model}}
In the first experiment, we use large vision-language models (LVLMs) to capture the semantic content of the frames from the museum videos. From each video, 32 frames were uniformly extracted, and their embeddings were computed. 
These features were then fed into the hierarchical method described in Section \ref{sec_method}. For this step, three models, including OpenClip \cite{radford2021learning}, Blip \cite{li2022blip}, and MobileClip \cite{vasu2024mobileclip}, were employed to obtain embeddings for both the video frames and the museum descriptions using the pretrained visual and textual encoders of the three models, respectively. The results are presented in Table \ref{tab:vision_language_model}. As shown, the best results are achieved using OpenClip as the Pretrained VL Knowledge component. While Blip also achieves notable results (e.g. obtaining the same R@1 and only about 3\% lower R@5 than OpenClip), this experiment confirms the importance of carefully choosing the large vision-language model used to implement the Pretrained VL Knowledge component.

\begin{table}[]
    \centering
    \caption{Comparison using different vision-language models. Details are explained in Section \ref{subsec_vision_language_model}.}
    \begin{tabular}{|c|cccccc|} \hline
        \multicolumn{1}{|c|}{} & \multicolumn{3}{c}{Recall} & & & \\
        Model & R@1 & R@5 & R@10 & MedR & MeanR & MRR \\
        \hline
         OpenClip &\textbf{62.31} &\textbf{97.10} &\textbf{100} &\textbf{1} &\textbf{1.79} &76.42 \\
         Mobile-Clip &44.92 &84.05 &92.75 &2 &3.47 &55.44  \\
         Blip &\textbf{62.31} &94.20 &97.10 &\textbf{1} &2.08 &\textbf{77.92}  \\
        \hline
    \end{tabular}
    \label{tab:vision_language_model}
\end{table}

\subsection{Impact of large video-only models\label{subsec_video_models}}
Since AgriMuseums mainly feature videos, we examine whether large video models yield better representations than vision-language models that treat videos as separate frames.
In this experiment, we use ViViT \cite{arnab2021vivit}, VideoMAE \cite{tong2022videomae}, and Clip4Clip \cite{luo2022clip4clip}. For the ViViT and Clip4Clip models, 32 frames were uniformly sampled from each video, while for VideoMAE, 16 frames were used due to its architectural requirements. To experiment with these, we adapt our method to accommodate the characteristics of the video-only models. Specifically, the In-Domain Adapter component of the Video Encoding module is removed, since video-only models natively aggregate the frames into a video-level representation. Thus, the extracted video-level features are fed directly into the Room Encoder, skipping the first level of the hierarchical pipeline. Moreover, while Clip4Clip supports both visual and textual modalities, ViViT and VideoMAE are video-only models and thus do not include a text encoder. For these, we use OpenClip textual encoder to extract textual features. 


The results of this experiment are presented in Table \ref{tab:video_models}, highlighting the superior performance of the ViViT model compared to the others. However, its performance is slightly lower than that of the vision-language models presented in Table \ref{tab:vision_language_model}, which may be due to the domain gap between the visual features and the textual ones. Nonetheless, video models tend to be more expensive than their image-based counterpart, e.g., at least about 284 GFLOPs for ViViT compared to about 20 GFLOPs for Clip with a ViT-B, making vision-language models more preferable.

\begin{table}[]
    \centering
    \caption{Comparison using different large models intended for video processing. Details are explained in Section \ref{subsec_video_models}.}
    \begin{tabular}{|c|cccccc|} \hline
        \multicolumn{1}{|c|}{} & \multicolumn{3}{c}{Recall} & & & \\
        Model & R@1 & R@5 & R@10 & MedR & MeanR & MRR \\
        \hline
         ViVit &\textbf{65.21} &\textbf{95.65} &\textbf{98.55} &\textbf{1} &\textbf{1.82} &\textbf{78.51} \\
         VideoMAE &47.82 &91.30 &94.20 &2 &2.79 &69.93 \\
         Clip4Clip &37.68 &82.60 &94.20 &2 &3.59 &44.13 \\
        \hline
    \end{tabular}
    \label{tab:video_models}
\end{table}

\subsection{Combining vision-language and video-only models \label{subsec_combination_models}}
As in the previous experiments, both video-only and vision-language models proved effective for the task of retrieving AgriMuseums, in this experiment, we aim at investigating whether combining them can lead to further improvements. We explore two types of fusion strategies: early fusion, where features are combined at the video level, and late fusion, where features are combined at the room level. We use OpenClip and ViViT models to represent the visual data as the two best models, while OpenClip features were used for the textual descriptions. 

The results are shown in Table \ref{tab:combination_models}, where the best results obtained in the previous sections are also included for easier comparison. As can be seen, Late Fusion has a better overall performance, achieving up to 63.78\% R@1 and 79.56\% MRR compared to 60.86\% R@1 and 73.26\% MRR obtained by Early Fusion. However, compared to using the single models, combining the models does not prove to be very effective, since ViViT alone achieves better R@1 (65.21\%) and OpenClip better R@5 (97.10\%). Nonetheless, Late Fusion leads to improved MRR (+1.05\%) compared to the single models.

\begin{table}[]
    \centering
    \caption{Comparison between two approaches (Late and Early Fusion) combining video-only and vision-language models for the Pretrained VL Knowledge component. Details are explained in Section \ref{subsec_combination_models}.}
    \begin{tabular}{|c|cccccc|} \hline
        \multicolumn{1}{|c|}{} & \multicolumn{3}{c}{Recall} & & & \\
        Model & R@1 & R@5 & R@10 & MedR & MeanR & MRR \\
        \hline
         Early Fusion &60.86 &94.20 &\textbf{100} &\textbf{1} &1.91 &73.26 \\
         Late Fusion &\underline{63.78} &\underline{95.65} &\underline{98.55} &\textbf{1} &1.88 &\textbf{79.56}\\
        \hline
        \textcolor{gray}{OpenClip} &\textcolor{gray}{62.31} &\textcolor{gray}{\textbf{97.10}} &\textcolor{gray}{\textbf{100}} &\textcolor{gray}{\textbf{1}} &\textcolor{gray}{\textbf{1.79}} &\textcolor{gray}{76.42} \\
        \textcolor{gray}{ViVit} &\textcolor{gray}{\textbf{65.21}} &\textcolor{gray}{\underline{95.65}} &\textcolor{gray}{\underline{98.55}} &\textcolor{gray}{\textbf{1}} &\textcolor{gray}{\underline{1.82}} &\textcolor{gray}{\underline{78.51}} \\
        \hline    
    \end{tabular}
    \label{tab:combination_models}
\end{table}

\subsection{Assessing the zero-shot performance of LVLMs \label{subsec_zero_shot}}
In the cross-modal retrieval literature, LVLMs such as Clip are often tested under zero-shot settings to validate their strong generalization capabilities. Here, we aim at exploring this type of approach with the AgriMuseums that we collected. We considered the AgriMuseums included in the test set and, based on the work presented in our work \cite{abdari2025agrimus}, selected the best combinations of aggregation functions at all three levels, including frames, videos, and rooms, to represent the visual content of each museum. For the textual features, we compute the average of the sentence embeddings for each museum description. Previous experiments showed that OpenClip features performed better; thus, the OpenClip model was used to represent both visual and textual content. Also, we consider Clip4Clip to perform a similar experiment on video-based models, as it integrates a textual encoder. The results, presented in Table \ref{tab:zero_shot}, indicate that using Mean aggregation for all three levels leads to the best performance. Nonetheless, while in existing works the zero-shot approach could lead to decent performance (up to 27\% R@1) \cite{abdari2025agrimus}, the performance observed on our dataset is far lower (up to about 3\% R@1). This considerable difference, especially when compared to that of the trained model (about 62\% R@1), is possibly due to the different textual data used in the two scenarios, as longer and more detailed descriptions are used here, and not only topics. 

\begin{table}[]
    \centering
    \caption{Comparison of Zero-shot Approaches. Details are explained in Section \ref{subsec_zero_shot}}
    \begin{tabular}{|c|ccc|cccccc|} \hline
        \multicolumn{1}{|c|}{} & \multicolumn{3}{|c|}{Aggregation of} & \multicolumn{3}{c}{Recall} & & & \\
        Model & Frames & Videos & Rooms  & R@1 & R@5 & R@10 & MedR & MeanR & MRR \\
        \hline
         & Mean & Mean & Mean &2.89 &13.04 &23.18 &22 &26.6 &10.2 \\
        Zero Shot &Median &Mean & Mean &1.44 &5.79 &15.21 &45 &53.23 &6.02 \\
        Image Model&Mean &Median &Mean &0.96 &4.34 &9.17 &69 &79.47 &4.45 \\
         &Median &Mean &Mean &0.72 &3.26 &7.24 &90 &106.41 &3.55 \\
        \hline
        Zero Shot &- &Mean & Mean &1.44 &7.24 &15.94 &39 &36.84 &6.75 \\
        Video Model &- &Median &Mean &0.72 &4.34 &6.52 &75 &73.40 &3.89 \\
        \hline
         Ours &\multicolumn{3}{|c|}{Trained Aggregation} &\textbf{62.31} &\textbf{97.10} &\textbf{100} &\textbf{1} &\textbf{1.79} &\textbf{76.42}  \\
        \hline
    \end{tabular}
    \label{tab:zero_shot}
\end{table}

\subsection{Ablation on the hierarchical structure \label{subsec_non_hierarchical}}
As the proposed approach is based on a hierarchical structure, in this experiment we ablate it to validate its effectiveness when used in its entirety. In this setting, we consider a total of three additional models. First, a flat model, shown in Table as \texttt{NHL(f\_{museum})}, is used to represent the museum by combining the visual content directly, i.e., ignoring the In-Domain Adapter and the Room Encoder. 
Then, the other two either ignores the Room Encoder (\texttt{NHL($f_{video}$ + $f_{museum}$)}) or the In-Domain Adapter (\texttt{NHL($f_{room}$ + $f_{museum}$)}). 
As in previous cases, the Pretrained VL Knowledge is implemented through the OpenClip model. 

The results in Table \ref{tab:non_hierarchical} confirm the effectiveness of modeling the data across all the three levels, as removing either of the components results in far lower performance (e.g., dropping between 14.49\% and 26.08\% R@1). 

\begin{table}[]
    \centering
    \caption{Ablation study of the hierarchical structure leveraged by our approach. Here, HL stands for Hierarchical and NHL stands for Non-Hierarchical Learning. Details are explained in Section \ref{subsec_non_hierarchical}.}
    \begin{tabular}{|c|cccccc|} \hline
        \multicolumn{1}{|c|}{} & \multicolumn{3}{c}{Recall} & & & \\
        Model &R@1 &R@5 &R@10 &MedR &MeanR &MRR \\
        \hline
        Zero Shot (H $f_{\text{video}} + f_{\text{room}} + f_{\text{museum}}$) &2.89 &13.04 &23.18 &22 &26.6 &10.2 \\
        \hline
        NHL($f_{\text{museum}}$) &37.68 &85.50 &95.65 &2 &3.42 &57.33 \\
        NHL($f_{\text{video}} + f_{\text{museum}}$) &47.82 &82.60 &97.10 &2 &3.18 &53.72 \\
        NHL($f_{\text{room}} + f_{\text{museum}}$) &36.23 &86.95 &97.10 &2 &2.94 &59.25 \\
        \hline
         Ours (HL $f_{\text{video}} + f_{\text{room}} + f_{\text{museum}}$) &\textbf{62.31} &\textbf{97.10} &\textbf{100} &\textbf{1} &\textbf{1.79} &\textbf{76.42} \\
        \hline
    \end{tabular}
    \label{tab:non_hierarchical}
\end{table}

\subsection{Transfer to external benchmarks \label{subsec_ircdl_test}}
Finally, we evaluated our model, which was trained on the dataset collected in this work using OpenClip features, on the dataset collected in \cite{abdari2025agrimus}. 
This dataset contains 83 museums structurally similar to ours, although they lack textual descriptions, as each museum is paired to a simple annotation only containing the topic related to all its rooms. To have a fairer comparison, since our model is trained using longer descriptions, we created brief textual descriptions for each of the museums and used those as queries. For instance, for a museum with 4 rooms, the description starts with "In this museum there are four rooms". Then, it follows with the number of videos and the subject of the room, like "In the first room, there are four videos about plant potato". The obtained results are presented in Table \ref{tab:ircdl_comparison}. Even though the descriptions are different and this time, are more complex and longer, as can be seen, the performance has improved by about 6.5\% R@1 and 11.23\% MRR. 

\begin{table}[]
    \centering
    \caption{Evaluation of the proposed approach, trained on the dataset collected in this work, on the dataset of \cite{abdari2025agrimus}. Details are explained in Section \ref{subsec_ircdl_test}.}
    \begin{tabular}{|c|cccccc|} \hline
        \multicolumn{1}{|c|}{} & \multicolumn{3}{c}{Recall} & & & \\
        Model &R@1 &R@5 &R@10 &MedR &MeanR &MRR \\
        \hline
        Zero Shot \cite{abdari2025agrimus}  &27.23 &56.33 &75.58 &4 &- &41.33 \\
        \hline
         Ours &\textbf{33.73} &\textbf{68.67} &\textbf{75.90} &\textbf{2} &8.04 &\textbf{52.56}  \\
        \hline
    \end{tabular}
    \label{tab:ircdl_comparison}
\end{table}

\section{Limitations and Future Work \label{sec_limitations}}
Several limitations and potential directions for future work can be identified. First, there is a need to create a larger and more diverse dataset covering a wider range of agricultural topics. Since obtaining a sufficient amount of high-quality data can be challenging, expanding the dataset is a key area for future work that could significantly enhance the overall performance of the system. This could be done both by integrating videos about other topics and by introducing 3D objects and interactive experiences, such as pruning virtual trees, to further promote the educational aspect.

Another promising direction is the incorporation of more natural and user-like textual queries. By developing a query bank that reflects the way ordinary users phrase their searches and using it during the training phase, the system's ability to retrieve relevant museums based on natural inputs can be greatly improved.

Lastly, conducting a comprehensive user study is essential. This would help identify shortcomings and nuanced issues from the end users' perspective, providing valuable insights to refine and improve the system as a whole.   

\section{Conclusion \label{sec_conclusion}}
In this work, we focus on the problem of retrieving Agricultural Museums through textual queries. Specifically, as existing benchmarks for this task are quite small, we created a new dataset consisting of 457 thematic museums, each covering a specific topic within the agricultural sector. In addition, we implemented and trained a retrieval system based on a hierarchical vision-language approach jointly modeling both the Agricultural Museums and the textual descriptions. In our experimental setup, we explored the integration of different image-based and video-based models, and obtained up to 62.31\% R@1 and 76.42\% MRR using OpenClip features. The ablation studies confirmed the effectiveness of each level in the hierarchical structure, which if removed led to considerable drops (e.g., about -15\% R@1). Moreover, training the model on the collected datasets led to better performance on existing benchmarks (e.g., +11\% MRR).

\bibliographystyle{splncs04}
\bibliography{main}
\end{document}